\def\year{2020}\relax
\documentclass[letterpaper]{article} 
\usepackage{aaai20}  
\usepackage{times}  
\usepackage{helvet} 
\usepackage{courier}  
\usepackage[hyphens]{url}  
\usepackage{graphicx} 
\usepackage{times}
\usepackage{epsfig}
\usepackage{graphicx}
\usepackage{amsmath}
\usepackage{amssymb}
\usepackage{float}
\usepackage{fancyhdr}

\usepackage{algorithm}
\usepackage{algorithmic}

\usepackage{graphicx}  
\title{Active Learning in Video Tracking}
\author{ Sima Behpour\\
University of Pennsylvania \\
sbehpour@seas.upenn.edu \
}
 
\begin{document}

\maketitle

\begin{abstract}
Active learning methods, like uncertainty sampling, combined with probabilistic prediction techniques have achieved success in various problems like image classification and text classification. For more complex multivariate prediction tasks, the relationships between labels play an important role in designing structured classifiers with better performance. However, computational time complexity limits prevalent probabilistic methods from effectively supporting active learning. Specifically, while non-probabilistic  methods based on structured support vector machines can be tractably applied to predicting bipartite matchings, conditional random fields are intractable for these structures. We propose an adversarial approach for active learning with structured prediction domains that is tractable for matching. We evaluate this approach algorithmically in an important structured prediction problems: object tracking in videos. We demonstrate better accuracy and computational efficiency for our proposed method.
\end{abstract}
\section{Introduction}
\begin{figure}[t]
\centering
    \includegraphics[width= 0.99\columnwidth]{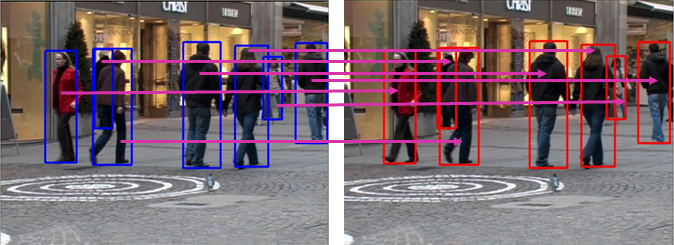}
    \caption{An example of a bipartite matching in a video tracking application. When some assignments are uncertain, choosing the most informative one can significantly reduce any uncertainty in the remaining assignments.}
    \label{fig:videoTracking}
    \vspace{-3mm}
\end{figure}

\noindent In many real-world applications, obtaining labeled instances for training is expensive.
This is particularly true for multivariate prediction tasks, in which many labels are required for each training example.
For example, an image can require many tags (e.g., mountain, sky, tree) as part of a multi-label prediction task, and video tracking has many pairs of bounding boxes between consecutive frames (Figure \ref{fig:videoTracking}).

Exhaustively annotating datasets for these tasks is extremely burdensome. Active learning \cite{settles2012active} seeks to reduce this annotation burden by only requesting the annotations that are most useful for learning. In \cite{sinha2019variational}, a pool-based semi-supervised active learning algorithm is designed that implicitly learns
the sampling mechanism in an adversarial manner. 

In this paper, we specifically consider the multivariate active learning setting. Many important structured prediction problems, including learning to rank items, correspondence-based natural language processing, and multi-object tracking,
can be formulated as weighted bipartite matching optimizations. 
Existing structured prediction approaches have significant drawbacks when applied under the constraints of perfect bipartite matchings.
Exponential family probabilistic models, such as the conditional random field (CRF), provide statistical consistency guarantees, but suffer computationally from the need to compute the normalization term of its distribution over matchings, which is a \#P-hard matrix permanent computation.  
In contrast, the structured support vector machine (SSVM) provides computational efficiency, but lacks Fisher consistency, meaning that there are distributions of data for which it cannot learn the optimal matching even under ideal learning conditions (i.e., given the true distribution and selecting from all measurable potential functions). 
We propose adversarial bipartite matching to avoid both of these limitations and employ an efficient active learning strategy for this framework.
The paper is organized as follows. Firstly, we introduce background and related work, which includes the main methodology of adversarial structured prediction \cite{asif2015adversarial,fathony2018efficient}--- Adversarial Bipartite Matching (ABM) method, followed by the active learning algorithm details. We compare the proposed method with state-of-the-art approaches on
real-world data sets in experiments before we conclude.
\section{Background and Related Works}
\subsection{Multivariate Active Learning}
We consider multivariate active learning in this work. Instead of a uni-variate label, $y$, each example has a vector-valued label ${\bf y}$. 

We consider the scenario where the active learner can solicit single variables in this label vector,
$y_i$, instead of soliciting the entire vector, ${\bf y}$, at each iteration of learning.

In multivariate classification \cite{tsoumakas2007multi}, each instance $x \in R ^d$ is mapped to a label-set $y \subseteq \{1,2,3,...,N\}$, where N is the number of classes. The label-set y is presented as a N dimensional binary vector in $\{0, 1\}^N$, where the n-th element is 1 if the instance belongs to th class label n. Assume $D=\{(x_1,y_1),(x_2,y_2),...,(x_m,y_m)\}$ is the data with m training examples, a multivariate classifier learns a function $f:R^d\rightarrow \{0,1\}^N$ from $D$.
We consider pool-based active learning \cite{lewis1994sequential} setup for multivariate classification in this work. In this setting, $D_l$ (labeled pool) represents the labeled training sets and $D_u$ presents an unlabeled pool. The algorithm runs R rounds of queries. In every round, the active learning algorithm calls the learner F on labeled pool ($D_l$) and learns a function $f$. Based on selection strategy and $\{f, D_l,D_u\}$, it first selects a subset of unlabeled data from $D_u$ and then chooses a subset of class labels (N classes) for them to be queried. It is notable that the full label-vector $y$ may not be solicited and it may query a subset of N labels for an instance. The new labeled data are added to $D_l$ and removed from $D_u$. The algorithm goal is to learn the best classifier with the minimum number of labeled instances and classes. 

The simplest approaches to multivariate active learning reduce each example to a set of uni-variate examples and apply uni-variate active learning methods. 


\subsection{Adversarial Structured Prediction}

Many   important   structured   prediction   problems like scheduling problems and multi-object tracking can be modeled as  weighted  bipartite  matching  problem. Adversarial Bipartite Matching provides fisher consistency and efficient time complexity.Having these two important properties give privileges to this method in compare with SSVM and exponential models suffering from lack of fisher consistency or high time complexity. The ABM approach  is  based on an adversarial classification \cite{topsoe1979information,grunwald2004game,asif2015adversarial}. It models the learning phase as a data-constrained zero-sum game between an adversary seeking to maximize the expected loss and a predictor seeking to minimize the expected loss. 
It trains a predictor that robustly minimizes the Hamming loss against the worst-case permutation mixture probability which is consistent with the training data statistics. In this zero-sum game, there are two players: a predictor and an adversary. The predictor makes a probabilistic prediction( $\hat{p}$) over the set of all possible assignments against an adversary making probabilistic prediction ($\check{p}$) matching the training data statistics ($\tilde{p}$).
\subsection{Minimax Game Formulation}

Rather than seeking a predictor that minimizes the (regularized) empirical risk,
\begin{align}
    \min_\theta \mathbb{E}_{}\left[\text{loss}(Y,f_\theta(X))\right] + \lambda ||\theta||_2,
\end{align}
adversarial prediction methods \cite{topsoe1979information,grunwald2004game} instead introduce an adversarial approximation of the training data labels, $\check{P}(\check{y}|x)$, and seek a predictor, $\hat{P}(\hat{y}|x)$, that minimizes the expected loss against the worst-case distribution chosen by the adversary:
\begin{align}&\quad \min_{\hat{P}} \max_{\check{P}} \mathbb{E}_{x \sim \tilde{P}; \check{y}|x \sim \check{P}; \hat{y}|x \sim \hat{P}}\left[ \text{loss}(\hat{Y},\check{Y})\right] \\
    & \text{such that: }\mathbb{E}_{x \sim \tilde{P}; \check{y}|x \sim \check{P}}\left[ \phi(X,\check{Y})\right] = \tilde{\bf c},\notag\end{align}
where the adversary is constrained by certain measured statistics (i.e., based on feature function $\phi$) of the training sample $\tilde{\bf c}$---either with equality constraints, as shown, or inequality constraints. $\tilde{P}$ represents the empirical distribution of ${\bf X}$ [and ${\bf Y}$] in the training data set.

While the empirical risk cannot be tractably optimized for many natural loss functions of interest (e.g., the 0-1 loss or Hamming loss), adversarially minimizing the loss measure is often tractable \cite{asif2015adversarial}.  
This adversarial minimization aligns the training objective to the loss measure better than surrogate losses (e.g., the hinge loss), providing better performance in practice for both classification \cite{asif2015adversarial} and structured prediction \cite{BehpourXZ18,fathony2018efficient}

A key advantage of this adversarial approach for multivariate active learning is that the adversary chooses a joint probability distribution, which provides correlations between unknown label variables, $P(y_i,y_j)$, that are useful for estimating the value of information for different annotation solicitation decisions.
The benefit of this uncertainty in structured predictions is most pronounced for settings in which other probabilistic methods---namely, conditional random fields \cite{lafferty2001conditional}---are computationally intractable, while adversarial structured prediction methods can be efficiently employed. We focus on active learning for two such structured prediction tasks in this paper: learning to make cuts in graphs and learning to make bipartite matchings. In the remainder of this section, we review the adversarial structured prediction methods for these settings.
\subsection{Adversarial Bipartite Matching (ABM)} 
\begin{figure}[H]
\centering
    \includegraphics[width= 1.0\columnwidth]{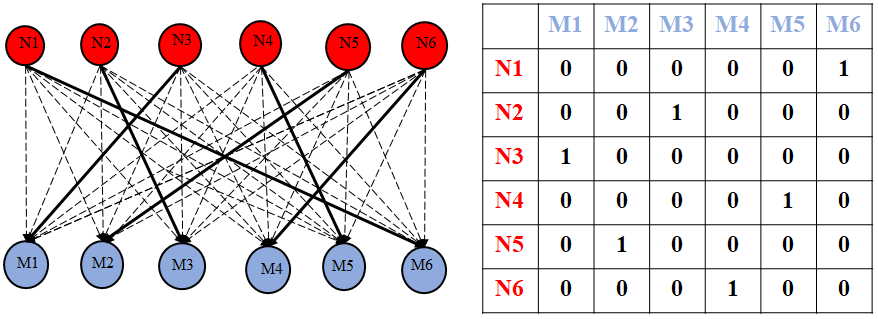}
    \caption{A bipartite matching graph of size six, $|M|=|N|=6$, and the adjacency matrix.}
    \vskip -2mm
    \label{fig:bpgraph}
\end{figure}

A bipartite graph G = (V, E) is a graph that its vertex set V can be divided into two same size disjoint nodes groups M and N provided every edge $ e \in E$ has one end point in M and the other end point in N.
A perfect bipartite matching B seeks a subset of edges such that each node in V appears in at most one edge of the bipartite graph, , as presented in Figure \ref{fig:bpgraph}.
This matching 
can be denoted as an assignment (or permutation) $\pi$,
where $\pi_i \in [n]$ indicates the item in the second set that item $i$ from the first set is paired with.
ABM \cite{fathony2018efficient} seeks the predictive distribution over assignments with the lowest expected loss against a constrained adversarial approximation of the training data assignments:
\begin{align}
\label{eq:adv}
    & 
    \min_{\hat{P}(\hat{\pi}|x)}
    \max_{\check{P}(\check{\pi}|x)}
    \mathbb{E}_{x\sim \tilde{P};\hat{\pi}|x\sim \hat{P};
    \check{\pi}|x\sim \check{P}}\left[ \sum_{i=1}^n \hat{\pi}_i \neq \check{\pi}_i \right]
    \\
     & \text{such that } 
    \mathbb{E}_{x\sim \tilde{P};\check{\pi}|x\sim \check{P}}\left[\sum_{i=1}^n\phi_i(x,\check{\pi}_i) \right] \!= \tilde{\bf c},
\notag
\end{align}
where $\hat{\pi},\check{\pi}$ are the node assignments chosen by the predictor and the adversary,  respectively. 
This approach also applies the double oracle method \cite{mcmahan2003planning} to generate the set of active constraints (rows and columns of the game matrix) supporting the game's equilibrium.
Best responses are obtained using the Hungarian matching algorithm, also known as the Kuhn-Munkres algorithm, to find the maximum-weight matchings in a bipartite graph in $O(|V|^3)$ time.

\section{Adversarial Multivariate Active Learning}
In this section, we explain our sample selection strategy and apply it in bipartite matching. The comprehensive version of this work address multi-label classification problems as well \cite{behpour2019active}.

\subsection{Sample Selection Strategy}

A label selection strategy that provides the most useful information for learning is needed.
The full impact of soliciting a label is the combination of what it reveals about other variables in the structured prediction and what influence updating the model parameters will have on all other variables.
Since the latter is very difficult to calculate exactly or even estimate loosely, we focus on the former.
The benefit of observing a variable can be measured using information theory.
The total expected reduction in uncertainty over all variables, $Y_1, \hdots, Y_n$, from observing a particular variable $Y_j$ given labeled dataset $D_l$ (referred to as $V_j$) is: 
\begin{align}
V_j = 
& \overbrace{\sum_{i=1}^n H(Y_i|D_l)}^{\substack{\text{uncertainty before}\\
\text{observing $y_j$}}} 
- 
\overbrace{\sum_{y_j \in \mathcal{Y}} P(y_j|D_l)
\sum_{i=1}^n H(Y_i|D_l,y_j)}^{\substack{\text{expected uncertainty}\\ \text{after observing $y_j$}}} 
\notag\\
& = \sum_{i=1}^n I(Y_i; Y_j|D_l). \label{eq:value}
\end{align}
These mutual information values can be effectively computed from the adversary's equilibrium distribution using the pairwise marginal probabilities of two variables, $\check{P}(y_i,y_j)$, which was our main motivation for employing adversarial prediction methods for problems that are intractable for other probabilistic structured prediction approaches.

\subsection{Active Learning Adversarial Bipartite Matching}
One of the advantages of the adversarial classifier is providing a meaningful probabilistic classification framework based on loss value and moment matching constraints over training data. These probabilistic outputs can also be useful in active learning solicitation. Following pool-based active learning approach \cite{lewis1994sequential}, we extend it in bipartite matching problem using Adversarial Bipartite Matching (ABM). The procedure is presented as Algorithm \ref{alg:active-abm}. In this setting, the algorithm first selects a subset of unlabeled data from $D_u$ and then chooses a subset of edges (N nodes assignments) to be queried. It is notable that the full assignment $\pi$ may not be solicited and only a subset of N nodes assignment for an instance may be queried. The new labeled data are added to $D_l$ and removed from $D_u$. The algorithm's goal is to learn the best classifier with the minimum number of labeled instances.

\begin{algorithm}[ht]
\caption{ Active Learning Algorithm for Adversarial Bipartite Matching (ABM).}
\label{alg:active-abm}
\begin{algorithmic}[1]
\REQUIRE{
Features $\{{\bf \phi}_{i}(\cdot)\}$;
 Initial parameters $\theta$; Initial labels ${\bf \pi}_{\text{initial}}$ \\
 $D_l$: a small set of initially labeled examples;\\
$D_u$: the pool of unlabeled data for active selection;} 
\STATE Train an initial ABM model $f$ on $D_l$;
         
            \REPEAT{
            \STATE Make predictions with classifier using parameters $\theta$ for all samples in $D_u$;
            \STATE Calculate $v_i(x)$ using Equation \ref{eq:value} for all nodes of samples $x \in D_u $;
            \STATE Select the most informative sample according to $X^* = \max_x \sum_i v(x)$; 
            \STATE Query the sample's node assignments; 
            \STATE Move $X^*$ from $D_u$ to $D_l$;}
            \STATE Train and update the ABM model $f$ on $D_l$; 
           
        \UNTIL stop criterion reached.\\
        
\STATE \textbf{return} final classifier parameters $\theta$.
\end{algorithmic}
\end{algorithm}
\section{Experiments}
\begin{figure*}[t!]
\setlength\tabcolsep{0cm}
\begin{center}
\begin{tabular}{cccc}
(a) ETH-BAHNHOF & (b) TUD-CAMPUS & (c) TUD-STADTMITTE & (d) ETH-SUN \\\includegraphics[width=0.25\textwidth]{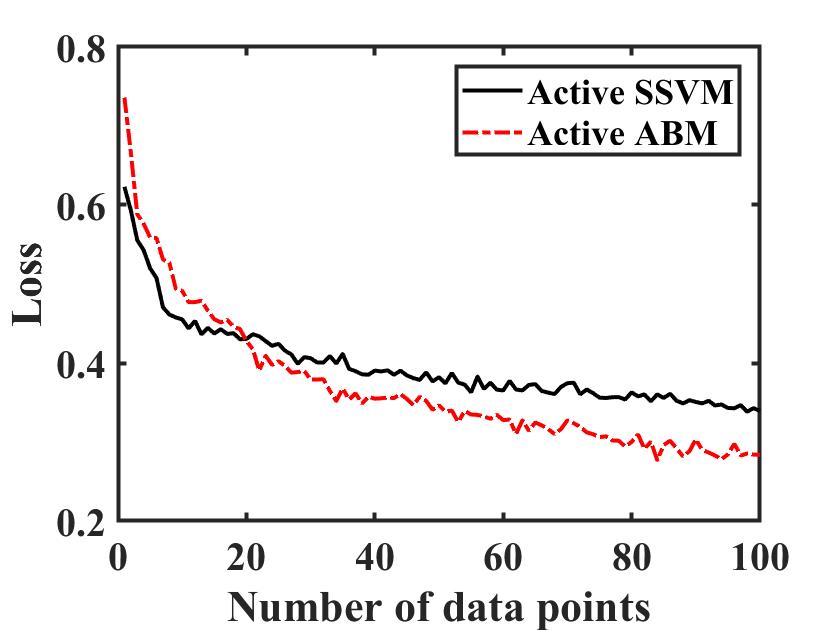}&
\includegraphics[width=0.25\textwidth]{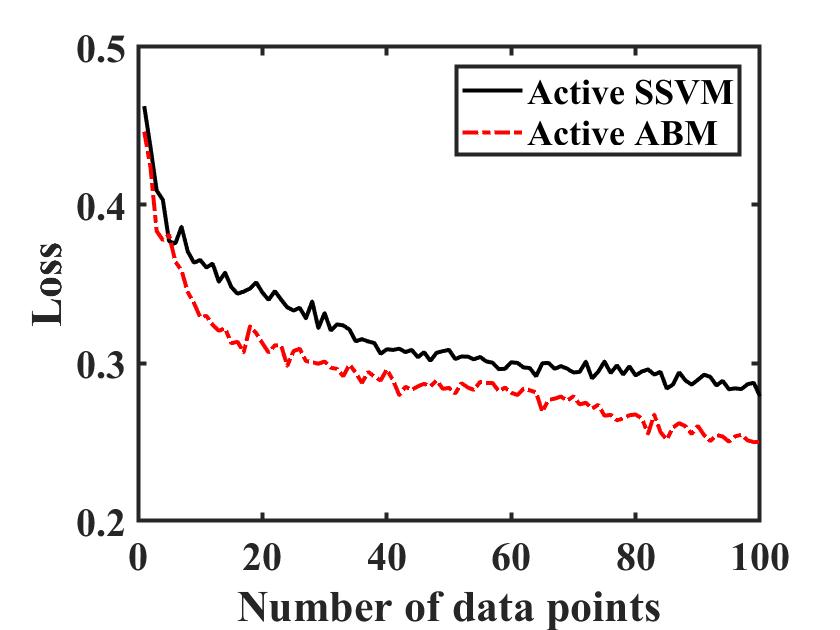}&
\includegraphics[width=0.25\textwidth]{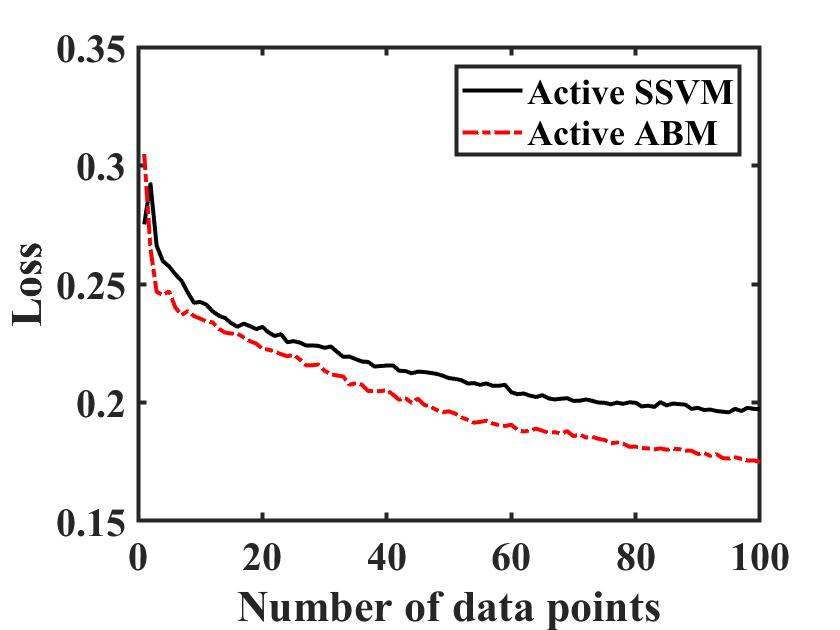}&
\includegraphics[width=0.25\textwidth]{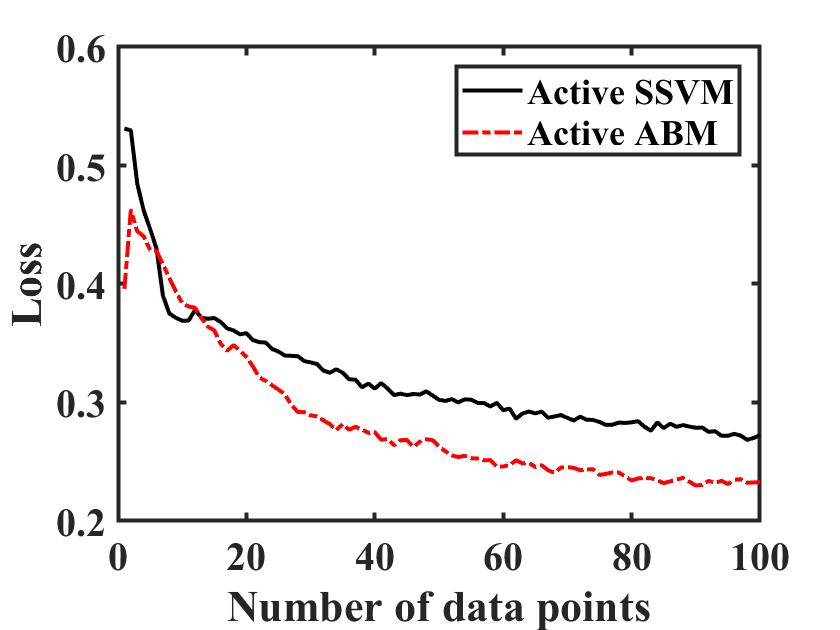}\\\\
(e) BAHNHOF/PEDCROSS2 & (f) CAMPUS/STAD & (g) SUN/PEDCROSS2 & (h) BAHNHOF/SUN \\\includegraphics[width=0.25\textwidth]{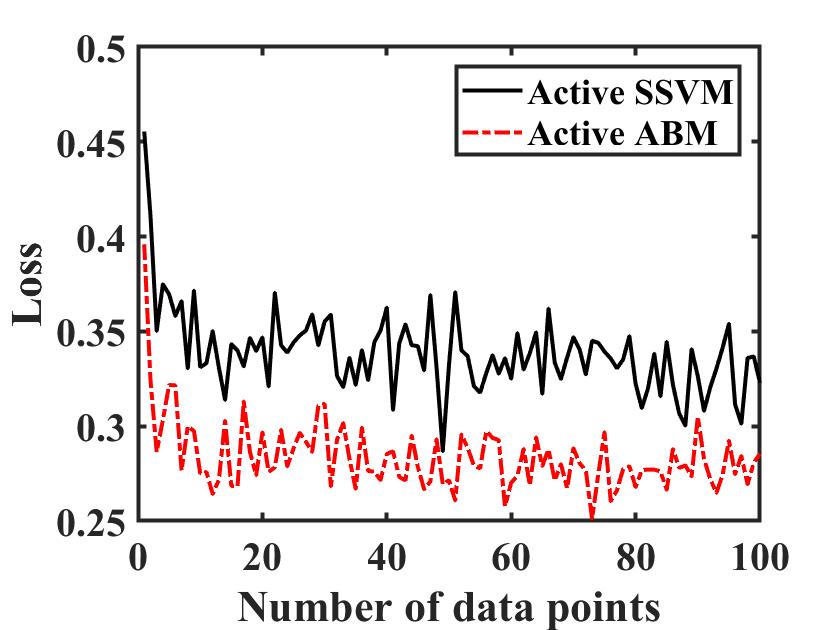}&
\includegraphics[width=0.25\textwidth]{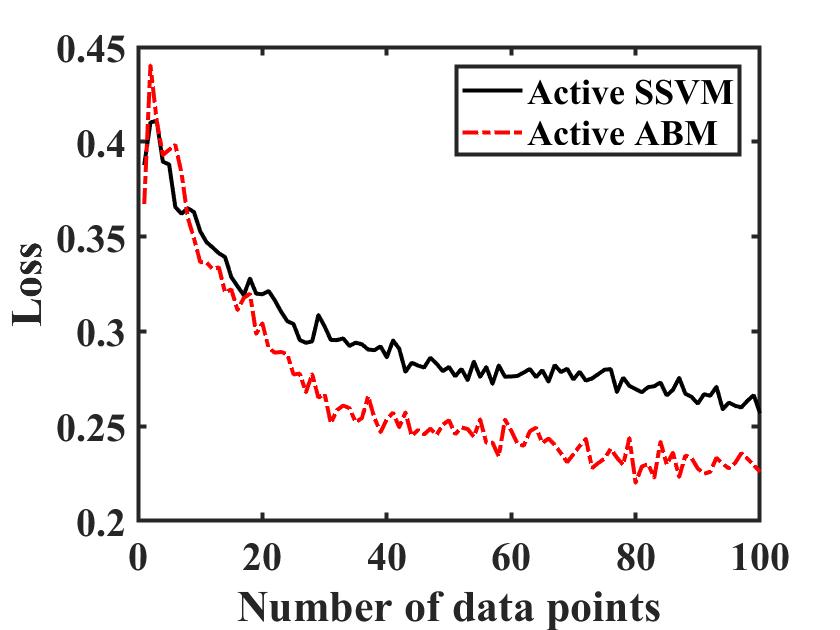}&
\includegraphics[width=0.25\textwidth,height=0.188\textwidth]{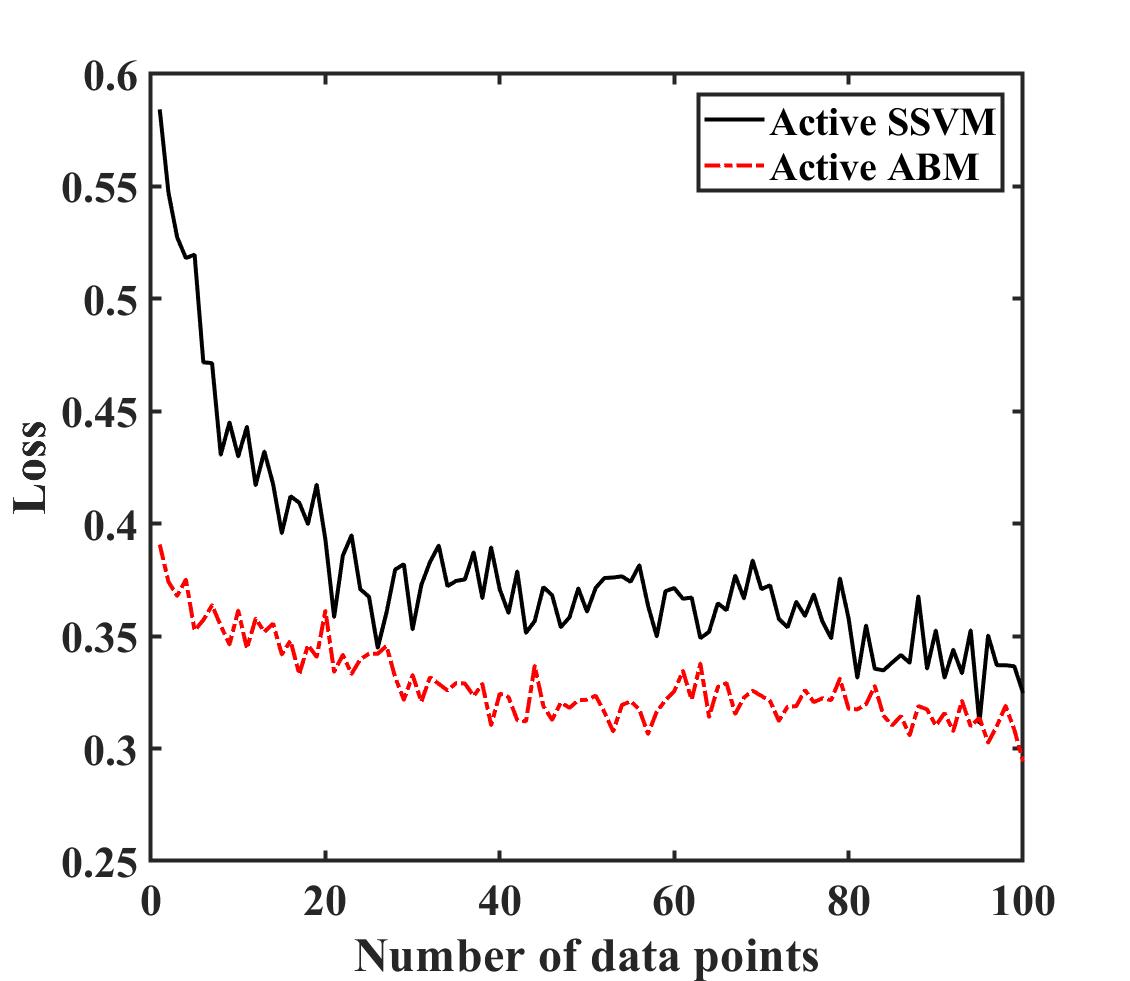}&
\includegraphics[width=0.25\textwidth]{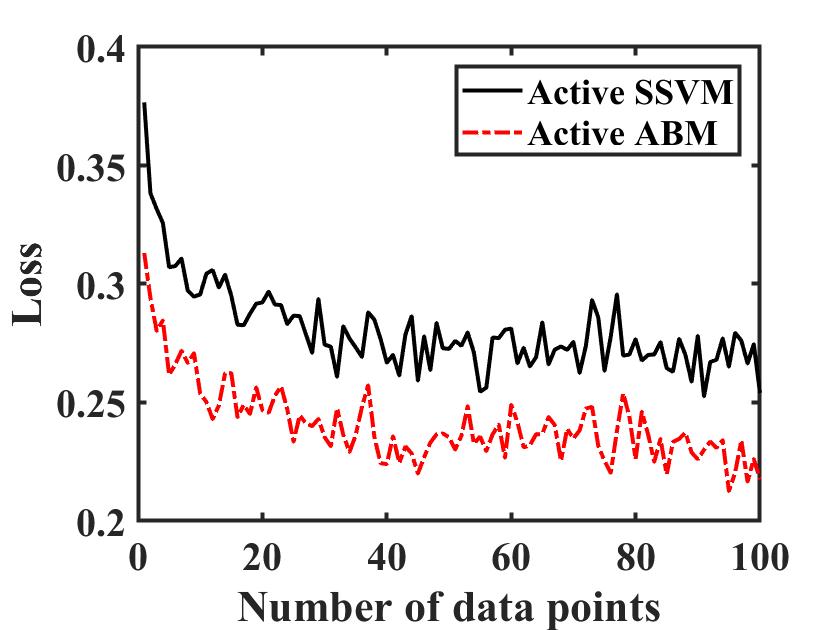}\\
\end{tabular}
\end{center}
\caption{Hamming loss (Loss) values for the first 100 data points of active learning averaged over 30 randomized withheld evaluation dataset splits.}
\label{fig:bpexp}
\end{figure*}
\subsection{Prediction Task, Datasets, and Features}
\paragraph{Object tracking:}
We consider active learning for object tracking between video frames.
The active learning selects two consecutive frames to add as labeled data to its training dataset to improve performance.
We follow the same problem definition presented in \cite{kim2012online}. In this setting, a set of images (video frames) and a list of objects in each image are given. There are many methods like ADA \cite{behpour2019ada} to detect objects in every video frame. The correspondence matching between objects in frame $t$ and objects in frame $t+1$ is also provided. 
Figure \ref{fig:videoTracking} provides an example of this setup. 
Considering the video frames, we can expect that the number of objects is changed in every frame. The number of objects changes when a subset of the objects may enter, leave, or remain in the consecutive frames. To deal with this problem, we double the number of nodes in every frame. We consider the number of objects in frame $t$ as $N_t$ and $
N^*$ be the maximum number of objects a frame can have, i.e., $N^* = \max_{t \in T} N_t$.
Starting from $N^*$ nodes to present the objects, we consider $N^*$ more nodes as ``invisible" nodes to let new objects to enter and existing objects to leave.
Hence the total number of 
nodes in each frame doubles to $n=2N^*$. We follow the joint feature representation in ABM and \cite{kim2012online} to define the affinities and correlations between node pairs in two consecutive frames.
We evaluate using a video tracking problem from the MOT challenge dataset \cite{leal2015motchallenge}.
We consider the TUD datasets and the ETH datasets as two different groups of datasets in our experiment. Each dataset differs in
the number of samples (i.e., pairs of two consecutive frames to be matched) and the number of nodes (i.e., the number of
pedestrian bounding boxes in the frame plus the number of extra nodes to indicate entering or leaving). The detailed information of datasets is demonstrated in Table 1.
To make the experiment more challenging and to avoid having examples that are too similar in the training set, we combine each pair of datasets that have similar characteristics. In particular, this results in four mixed datasets that we evaluate.

\subsection{Object Tracking Results} 
\paragraph{Structured support vector machines with Platt scaling: }
For our bipartite matching setting, we consider active learning with structured support vector machines (SSVM) as a baseline to evaluate the performance of our approach. We implement the SSVM model \cite{taskar2005learning,tsochantaridis2005large} following \cite{kim2012online} using
SVM-Struct \cite{joachims1998text}. We apply Platt scaling \cite{platt1999probabilistic} to transform our potential functions to probabilistic outputs under the SSVM. It works by fitting a sigmoid function to the decision values for each class through the optimizing parameters \emph{a} and \emph{b} of a Sigmoid function (called the scoring algorithm) parameters, in the following expression:
$
    \frac{1}{(1+exp (a\cdot z+b))}$,
where \emph{z} is an input potential value. We first follow a learning algorithm on a subset of data to learn and fit \emph{a} and \emph{b} in a multi-class setting. In solicitation step of active learning, we compute the probabilistic value of the bipartite graph edges by passing the potential value of the edge to the sigmoid scoring function. The entropy of every node is calculated by summing over the entropy of its edges 
and the entropy of every training sample is computed by summing the nodes entropies. 
The sample and the node with highest entropy is chosen to be solicited.
 It is the same solicitation strategy as active learning ABM which is applied to query the most informative node assignment. 

\begin{table}[t!]
\vskip -0.10in
\label{tab:ds}
\caption{Dataset properties}
\vskip -0.01in
\begin{center}
\begin{small}
\begin{sc}
\begin{tabular}{ | l | c | c | c | c |}
\hline
Dataset & \# Elements & \# Examples \\\hline
TUD-Campus  & 12 & 70 \\ 
TUD-Stadtmitte & 16 & 178 \\
ETH-Sunnyday & 18 & 353 \\ 
ETH-Bahnhof & 34 & 999 \\ 
ETH-Pedcross2 & 30 & 836 \\\hline
\end{tabular}
\end{sc}
\end{small}
\end{center}
\end{table}

We report the results of our
object tracking experiments in Figure \ref{fig:bpexp}.
Apart from somewhat mixed performance in the early iterations of active learning,
our proposed active learning framework (Active ABM) provides better performance compare to Active SSVM.  We attribute this advantage to the better uncertainty model that our approach provides compared with the Platt scaling approach used by SSVM.
\subsection{Inference Running Time}The key difference for inference under our approach (and advantage when learning) is that it uses multiple permutations to construct an equilibrium. The (average) numbers of permutations (denoted as Perms in the table) to arrive at an equilibrium for different datasets are presented in Table \ref{tab:runtime} for both experiments. We can conclude from Table \ref{tab:runtime} that inference from scratch is roughly 6-20 times slower than SSVM and other methods that use a single permutation. During training, however, the strategies from the previous equilibria can be cached and reused, making training time comparable to other methods.
\begin{table}[tb]
\vspace{-4mm}
\caption{Inference Running Time}
\label{tab:runtime}
\begin{center}
\begin{small}
\begin{sc}
\begin{tabular}{ | l | c | c | c | c | c |}
\hline
\hline
Dataset & \# Perms & Dataset &\# Perms \\\hline\hline
ETH-BAHN & 11.3 & TUD-CAMP  & 8.7 \\
TUD-STAD & 9.4 & ETH-SUN & 10.3\\
BAHN/PED2 & 8.3 & ECAMP/STAD & 10.9\\
SUN/PED2 & 15.1 & BAHN/SUN & 5.6\\
\hline
\end{tabular}
\end{sc}
\end{small}
\end{center}
\vspace{-5mm}
\end{table}
\section{Conclusion and Future work}
In this paper, we investigated active learning for an structured prediction task:learning bipartite matchings. 
Though structured support vector machines can be efficiently employed for these tasks, they are not very useful for guiding label solicitation strategies.
Conditional random fields, which do provide useful correlation estimates for computing value of information cannot be applied efficiently for these tasks (e.g., \#P-hard for bipartite matchings). 
We introduced active learning based on adversarial structured prediction methods that enjoy lower computational complexity than existing probabilistic methods while providing useful correlations between variables.
We demonstrated the benefits of this approach on an structured prediction task, object tracking, through consecutive video frames as a bipartite matching task.
For future work, we would like to extend our experiments by considering more datasets and investigate the efficiency of this method in different problems like object recognition, and semi-supervised classification.
\section{Acknowledgement}
I would like to thank Brian Ziebart for all his advise and support on this work.
{\small
\bibliographystyle{aaai}
\bibliography{egbib}
}

\end{document}